\documentclass[pdflatex,sn-mathphys-num,iicol]{sn-jnl}

\usepackage{graphicx}
\usepackage{amsmath}
\usepackage{amssymb}
\usepackage{amsfonts}
\usepackage{booktabs}
\usepackage{tabularx}
\usepackage{array}

\graphicspath{{figures/}{./}}

\begin{document}

\title[Self-Rewarded RL for Structured Medical Explanations]{Self-Rewarded Reinforcement Learning for Better-Structured Medical Explanations in Small Language Models}

\author[1]{\fnm{Prakhar} \sur{Mishra}}\email{443prakharmishra@gmail.com}

\author[2,4]{\fnm{Bhavya Yajush} \sur{Awasthi}}\email{bhavya.yajush@gmail.com}

\author[3]{\fnm{} \sur{Tanuja}}\email{shalus384@gmail.com}

\author*[1]{\fnm{} \sur{Nishant}}\email{nishantsinghchauhan0010@gmail.com}

\affil[1]{\orgdiv{Department of Information Technology},
  \orgname{Delhi Technological University},
  \orgaddress{\city{New Delhi}, \country{India}}}
  
\affil[2]{\orgname{Indian Institute of Technology, Guwahati},
    \orgaddress{\city{Guwahati}, \state{Assam}, \country{India}}}

\affil[3]{\orgname{Central University of Haryana},
  \orgaddress{\city{Mahendragarh}, \state{Haryana}, \country{India}}}

\affil[4]{\orgname{Delhi Skill and Entrepreneurship University}, 
    \orgaddress{\city{New Delhi}, \country{India}}}

\abstract{Small language models (SLMs) can run on commodity hardware and
support privacy-preserving, on-premises clinical deployment, but their medical
explanations are often short and poorly structured compared with large language
models. Recent supervised approaches such as Meerkat transfer chain-of-thought
reasoning into 7B--8B backbones, yet their training objective does not directly
target the style or organization of the generated rationales. Building on the
instruction-tuned Mistral-7B backbone, we investigate whether a lightweight
reinforcement learning (RL) step can improve explanation structure without
human preference labels. We apply Proximal Policy Optimization (PPO) with LoRA
adaptation to 2{,}000 Meerkat-derived medical prompts, guided by a simple
self-supervised reward that favors longer, lexically diverse, and
well-punctuated responses while penalizing very short ones. Training for about
170 PPO steps on a single NVIDIA A100 GPU yields a stable increase in the
average reward. Quantitative and qualitative comparisons show that the RL-tuned
model produces longer, more organized, multi-point explanations while
preserving the underlying medical content. We discuss the promise and the
limitations of self-rewarded RL for shaping the style of medical explanations.}


\maketitle

\section{Introduction}\label{sec:intro}

Large language models have shown great promise in question-answering in
medicine, case reasoning, and decision-making in the clinical domain. However,
the application of these models in the healthcare industry is limited due to
privacy and computational concerns.\cite{singhal2023largelms,nori2023gpt4medical}
Closed models like GPT-4 and Med-PaLM 2 can perform at or better than the level
of a human in USMLE tests, but they are not deployable in the cloud due to
confidentiality and compliance concerns.\cite{singhal2023largelms} On the other
hand, open SLMs with fewer than ten billion parameters can be run on a single
high-end GPU and are deployable in the healthcare
industry.\cite{kim2025meerkat,wu2023pmcllama,chen2023meditron} However, SLMs
have been observed to perform poorly in tasks that involve multi-step
reasoning, long context synthesis, and structured
explanation.\cite{kim2025meerkat,labrak2024biomistral}

The Meerkat series of medical SLMs have been designed to bridge this
performance gap by using chain-of-thought (CoT) reasoning traces from the GPT-4
model and transferring this reasoning ability into the 7B and 8B parameter
backbones Mistral-7B and Llama-3-8B, respectively, in combination with
synthetic QA pairs from 18 medical textbooks and several instruction following
datasets.\cite{kim2025meerkat} Fine-tuning the backbones of the SLMs on the
combination of CoT and instruction following data results in SLMs that perform
better than the existing 7B medical LMs and even the GPT-3.5 model on various
multiple-choice exams such as MedQA and MedMCQA while remaining deployable in
the on-premises environment.\cite{kim2025meerkat,jin2020medqa,pal2022medmcqa}
This demonstrates that carefully targeted supervision can endow SLMs with
strong reasoning skills without resorting to extremely large parameter counts.

In particular, reinforcement learning techniques such as Proximal Policy
Optimization (PPO)\cite{schulman2017ppo} have come to dominate the field of
alignment for LLMs with human preferences, as encapsulated by various instances
of the reinforcement learning from human feedback (RLHF) family of approaches
used for InstructGPT and similar
models.\cite{ouyang2022instructgpt,ziegler2019finetuning} While most existing
approaches in this space have focused on large models or used expensive
human/model-based preference labels for pairs of outputs, it remains an open
question for small medical language models: can a lightweight RL step, using a
simple self-supervised proxy reward, be used to improve the structure of the
generated explanations beyond what supervised CoT training can achieve.

This paper seeks to answer that question with an in-depth experimental design.
Beginning with the openly available Mistral-7B-Instruct
backbone,\cite{jiang2023mistral7b} we utilize PPO\cite{schulman2017ppo} with
LoRA adaptation\cite{hu2021lora} using the TRL library\cite{trl2023} on a
subset of Meerkat's CoT-style medical instruction data.\cite{kim2025meerkat}
The reward function is kept simple: it rewards longer outputs within a certain
length, lexical diversity, and a bonus if there is punctuation at the end of
sentences, while penalizing very short outputs. While such a reward does not
reflect clinical accuracy, it does directly affect explanation length and
structure, which is an important dimension in the limitations and future work
sections of Meerkat and similar medical
LMs.\cite{kim2025meerkat,wu2023pmcllama,chen2023meditron}

We summarize our contributions as follows:
\begin{enumerate}
  \item We propose a lightweight reinforcement learning pipeline for improving
  explanation structure in medical small language models using PPO with LoRA
  adaptation.

  \item We design a fully automatic proxy reward function based on response
  length, lexical diversity, and sentence boundary signals, enabling
  reinforcement learning alignment without human preference labels.

  \item Through both quantitative and qualitative analysis, we demonstrate that
  reinforcement learning can improve the structural organization of medical
  explanations generated by small language models.

  \item Our results highlight that reinforcement learning can be used not only
  to align language models with human preferences but also to control
  structural properties of generated explanations.
\end{enumerate}

\section{Background and related work}\label{sec:background}

\subsection{Medical small language models}\label{subsec:medslm}

There is an increasing amount of work on domain-specialized language models in
medicine and healthcare. Previous work has focused on fine-tuning
general-purpose LLMs using continuous pre-training on biomedical data such as
PubMed and clinical notes. This has led to the creation of models like
PMC-LLaMA, Me-LLaMA, and other variants of LLaMA specialized in
medicine.\cite{wu2023pmcllama,melama2024} While these models improve the
knowledge base of medical information, improvements in multi-step reasoning
capabilities are less impressive in proportion to the pre-training cost.

Meerkat proposes an alternate approach that does not rely on expensive
pre-training. Instead, it focuses on curating high-quality CoT datasets by
prompting GPT-4 on MedQA datasets and medical textbooks. Fine-tuning 7B and 8B
backbones on the combined CoT and instruction-following datasets results in
models that outperform previous 7B medical LMs like MediTron-7B, BioMistral-7B,
and GPT-3.5 on multiple-choice
exams.\cite{kim2025meerkat,chen2023meditron,labrak2024biomistral} This approach
shows that supervision over reasoning traces can narrow the performance gap
between medical LMs and even very large, expensive models like Med-PaLM and
Med-PaLM 2.\cite{singhal2023largelms}

Other medical LMs also investigate domain-specific adaptation: ChatDoctor,
MedInstruct-52K, and other instruction-tuned models focusing on
patient--doctor dialogue and consumer health-related
queries.\cite{chatdoctor2023,medinstruct2023} These studies emphasize the need
to consider the alignment of the output with the context of medical
communication beyond the results of the exam.

\subsection{Medical QA benchmarks}\label{subsec:benchmarks}

Multiple-choice medical question answer benchmarks like MedQA and MedMCQA have
become the standard in assessing the knowledge and reasoning abilities of LLMs
in the medical domain.\cite{jin2020medqa,pal2022medmcqa} MedQA is based on
USMLE questions in various languages from professional board exams, while
MedMCQA is based on almost 200k questions in 21 subjects from Indian entrance
exams.\cite{jin2020medqa,pal2022medmcqa} These two are the basis for the
evaluation protocols of Meerkat, Med-PaLM, PMC-LLaMA, MEDITRON, and
BioMistral.\cite{kim2025meerkat,singhal2023largelms,wu2023pmcllama,chen2023meditron,labrak2024biomistral}

Apart from the exam-oriented QA, the MultiMedQA system aggregates six existing
medical QA datasets with the HealthSearchQA dataset of consumer health queries
and presents a framework for human evaluation based on criteria like
factuality, precision, potential harm, and bias.\cite{singhal2023largelms} This
broader outlook again points out that success in multiple-choice tests does not
imply safe practice in the real world.

\subsection{Reinforcement learning for language models}\label{subsec:rl}

However, reinforcement learning has emerged as one of the main approaches to
align LLMs with human preferences and downstream objectives. Proximal Policy
Optimization (PPO) is one of the most popular on-policy policy gradient methods,
in which a clipped version of the surrogate objective is maximized in order to
ensure that the policy change remains in a `trust region' around the old
policy.\cite{schulman2017ppo} For LLMs, PPO is usually used in conjunction with
a reference model and a value function head, in which a reward is optimized that
consists of a scalar quality signal (such as a human preference score) as well
as a KL term that encourages the fine-tuned model to stay close to the
reference model.\cite{ziegler2019finetuning,ouyang2022instructgpt}

The RLHF paradigm that is instantiated by InstructGPT and subsequent models
uses PPO in addition to supervised fine-tuning, with a supervised reward
learning component that is trained on human preference
data.\cite{ziegler2019finetuning,ouyang2022instructgpt} More recent work has
focused on newer RL regimes that prioritize step-by-step reasoning
performance.\cite{wei2022cot,rlhf2023apa} However, existing studies focus on
larger general-purpose models.

A good implementation of PPO for transformer-based language models can be found
in the TRL library, including \texttt{AutoModelForCausalLM\allowbreak
WithValueHead} and \texttt{PPOTrainer} classes.\cite{trl2023} Together with
parameter-efficient adaptation approaches like LoRA, this makes it feasible for
researchers to test RL alignment approaches using their home computers.

\subsection{Parameter-efficient adaptation}\label{subsec:peft}

``Low-Rank Adaptation'' (LoRA) is a parameter-efficient fine-tuning method that
keeps the original weights of the model fixed and adds small matrices to be
learned to the linear layers that the model will be fine-tuned
on.\cite{hu2021lora} Instead of learning the entire weight matrix $W$ for the
linear layers, the weights $\Delta W$ will be learned in a low-rank factorized
form $BA$ where $A$ and $B$ have much lower ranks than the original weights.
This will result in a much lower number of parameters to be learned for the
model.

LoRA is part of a larger group of methods that are referred to as
``adapter-based.'' The idea behind these methods is that the backbone of the
network is frozen, and the adapters are small modules that are added for the
specific task at hand.\cite{houlsby2019adapter,pfeiffer2020adapterfusion} More
recently, there has been the introduction of QLoRA, where the idea is that the
combination of 4-bit quantization and the use of adapters for the task at hand
allows for the fine-tuning of large models (65B parameters) on a single GPU
without sacrificing performance.\cite{dettmers2023qlora} The combination of PPO
and LoRA for the task of RL for LLMs allows the policy and value functions to be
updated using only a small fraction of the total number of parameters, which is
particularly appealing for the case of 7B-scale models running on a single GPU.

\section{Method}\label{sec:method}

\subsection{Problem setting}\label{subsec:problem}

We assume an instruction-following single-turn medical context in which the
model is given a prompt in natural language $x$ that corresponds to a medical
query, symptom description, or conceptual query and is asked to produce an
answer $y$ that provides an explanation of relevant concepts and information,
possible reasons and explanations, and recommendations and actions to take.
Mathematically speaking, the language model is viewed as a policy
$\pi_\theta(y \mid x)$ over sequences of tokens, and the aim is to adjust the
parameters $\theta$ so that the produced explanation meets certain criteria in
terms of length, diversity, and sentence structure while retaining the learned
medical content from the instruction tuning.

Here, $f(y)$ is defined as the textual reward associated with the generated
answer based on surface-level statistics, as described below. Our objective is
to maximize the expected reward under the policy,
\begin{equation}
  J(\theta) = \mathbb{E}_{x \sim D,\, y \sim \pi_\theta(\cdot\mid x)}[f(y)],
\end{equation}
where $D$ is a fixed dataset of medical prompts derived from Meerkat's
instruction corpus.\cite{kim2025meerkat}

\subsection{Base model and dataset}\label{subsec:basemodel}

We load it from the \texttt{mistralai/\allowbreak Mistral-7B-\allowbreak
Instruct-v0.2} checkpoint, a 7B parameter decoder-only transformer with grouped
query attention and instruction tuning pretraining for general-purpose
dialogue.\cite{jiang2023mistral7b} This is used both as the reference and
initial policy. It has been pre-trained on web-scale data but has not been
fine-tuned with the specialized CoT supervision used in Meerkat or other medical
SLMs like PMC-LLaMA, MEDITRON, or
BioMistral.\cite{kim2025meerkat,wu2023pmcllama,chen2023meditron,labrak2024biomistral}

To add medical domain knowledge and structure to the reasoning, we fine-tune on
the JSONL dataset \texttt{meerkat\_\allowbreak rl\_\allowbreak subset.jsonl}
containing 2{,}000 prompts that were randomly selected from the instruction
tuning data released by Meerkat.\cite{kim2025meerkat} This is a self-contained
medical query or instruction that is usually in the form of medical questions,
USMLE questions, synthetic questions generated from textbooks, or
patient-doctor conversations. The variety of the sampled prompts includes
different types of medical instructions such as patient inquiries based on
symptoms, explanations based on concepts for physiological processes, medical
examination questions, and health advice scenarios. This diversity exposes the
model to both diagnostic reasoning and explanatory tasks during reinforcement
learning. For our initial experiment, we use PPO only on the prompts and do not
condition on the original Meerkat rationales, delegating explanation generation
entirely to the policy.

The tokenizer used for tokenizing the prompt is Mistral. It has a maximum input
length of 256 tokens and left padding. It generates up to 200 new tokens per
rollout in PPO training and evaluation.

\subsection{Reward design}\label{subsec:reward}

The reward function $f(y)$ is a deterministic function that takes the generated
answer text $y$ and returns a scalar value in approximately the range
$[0.01, 1.3]$. The reward function can be defined as a sum of four terms:
\begin{enumerate}
  \item \textbf{Length reward.} Let $n$ be the number of whitespace-separated
  tokens in $y$. We define
  \begin{equation}
    f_{\text{len}}(y) = \min\left(\frac{n}{50},\, 1\right),
  \end{equation}
  which increases linearly with length up to 50 tokens and saturates
  thereafter.

  \item \textbf{Diversity reward.} Let $V(y)$ be the set of unique tokens in
  $y$. We define the lexical diversity term as
  \begin{equation}
    f_{\text{div}}(y) = \frac{|V(y)|}{n + 10^{-6}},
  \end{equation}
  encouraging a higher ratio of unique words to total words.

  \item \textbf{Punctuation bonus.} To encourage explicit sentence boundaries,
  we add a fixed bonus if the text contains at least one sentence-ending
  punctuation mark:
  \begin{equation}
    f_{\text{punc}}(y) =
    \begin{cases}
      0.3, & \text{if } y \text{ contains any of } .,!?, \\
      0,   & \text{otherwise}.
    \end{cases}
  \end{equation}

  \item \textbf{Short-output penalty.} Extremely short responses are penalized
  via
  \begin{equation}
    f_{\text{short}}(y) =
    \begin{cases}
      -0.3, & n < 5, \\
      0,    & \text{otherwise}.
    \end{cases}
  \end{equation}
\end{enumerate}

The total reward is the sum
\begin{equation}
  f(y) = f_{\text{len}}(y) + f_{\text{div}}(y) + f_{\text{punc}}(y)
         + f_{\text{short}}(y),
\end{equation}
clamped from below at $0.01$ to avoid zero rewards. This proxy explicitly
encourages longer, lexically rich, and sentence-structured outputs while
disfavoring trivial or single-word responses. Notably, it does \emph{not}
encode any notion of medical correctness or safety, so the policy must rely on
prior supervised training for
factuality.\cite{ziegler2019finetuning,ouyang2022instructgpt}

\subsection{PPO fine-tuning with LoRA}\label{subsec:ppolora}

We perform the RL fine-tuning using PPO and the TRL library.\cite{trl2023} Both
the policy and the value function share the same backbone, which is based on the
Transformer architecture and implemented using
\texttt{AutoModelForCausalLM\allowbreak WithValueHead}. On top of this backbone,
we use a LoRA configuration that will be applied on the query, key, value, and
output projection matrices (\texttt{q\_proj}, \texttt{k\_proj},
\texttt{v\_proj}, \texttt{o\_proj}). We use a rank $r=16$, a scaling factor
$\alpha=32$, and a dropout probability of $0.05$.\cite{hu2021lora} All original
Mistral weights are frozen; only the LoRA adapters and value head parameters
are trainable.

The PPO is set with a batch size of 4 prompts per update, a minibatch size of
2, learning rate $10^{-6}$, initial KL coefficient 0.5, and target KL
divergence 3.0. The rollouts are generated by the current policy using nucleus
sampling ($\text{top}_p = 0.9$), temperature 0.7, repetition penalty 1.1, and
maximum 200 new tokens. The KL penalty is computed against a frozen reference
model, initialized from the same checkpoint before RL
training.\cite{ziegler2019finetuning,ouyang2022instructgpt}

The clipped PPO surrogate objective for a batch of trajectories can be written
as
\begin{equation}
  \begin{split}
    L^{\text{PPO}}(\theta) = \mathbb{E}_t\bigl[\min\bigl(
      &r_t(\theta)\,\hat{A}_t,\; \\
      &\operatorname{clip}(r_t(\theta), 1-\epsilon, 1+\epsilon)\,\hat{A}_t
    \bigr)\bigr],
  \end{split}
\end{equation}
where $r_t(\theta)$ is the ratio of the policy probability at the given token
position $t$ for the new and old policies, $\hat{A}_t$ is the advantage
estimate, and $\epsilon$ is the clipping parameter.\cite{schulman2017ppo} In our
case, the advantages are calculated for each token by the generalized advantage
estimation and are summed over the sequence.\cite{trl2023}

Training is run for 170--180 PPO steps, periodically saving checkpoints every 10
steps. All experiments are conducted on a single NVIDIA A100 GPU in a Google
Colab environment and all experiments were conducted using the Hugging Face
Transformers and TRL libraries with deterministic random seeds to ensure
reproducibility of training runs.

\section{Experiments}\label{sec:experiments}

\subsection{Training dynamics}\label{subsec:dynamics}

\begin{figure}[t]
  \centering
  \includegraphics[width=\columnwidth]{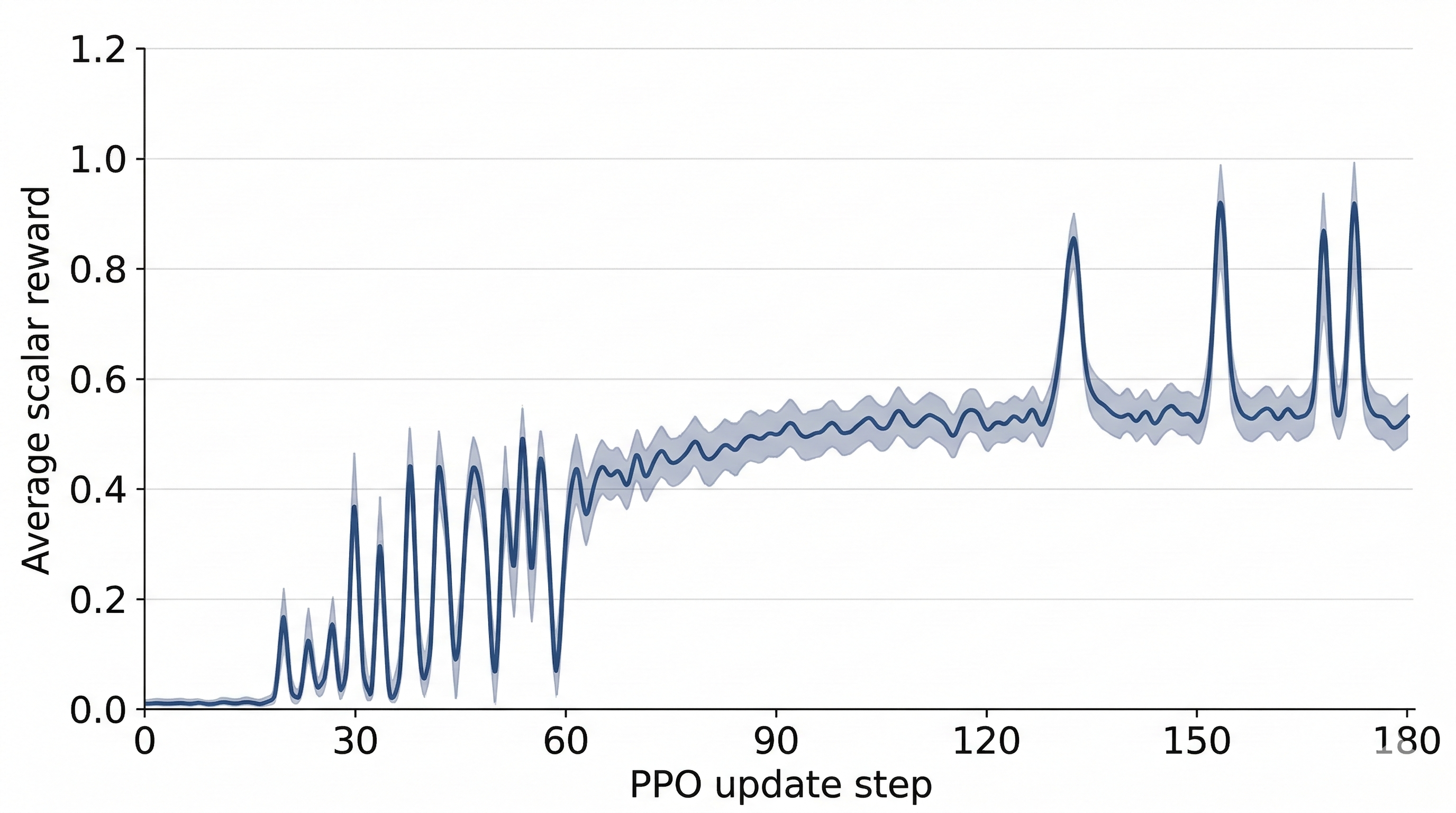}
  \caption{PPO training dynamics over the Meerkat-derived medical prompt
    subset. The plot shows the average scalar reward across more than 160 PPO
    updates. Initially, rewards cluster around the minimum clamped value of
    0.01, and later increase toward approximately 0.5 as the policy improves to
    generate longer and more varied outputs with appropriate punctuation under
    the proxy reward.}
  \label{fig:reward-trajectory}
\end{figure}

In the plot shown in Figure~\ref{fig:reward-trajectory}, the average scalar
reward over each PPO update cycle is tracked over 160+ updates. At the beginning
of the training process, the average reward is seen to vary between the minimum
clamp value of 0.01 and approximately 0.5. As training progresses, more updates
with high average reward are seen, with long sequences of updates yielding
average rewards between 0.45 and 0.55, with occasional spikes in reward up to
1.0 for long, diverse, well-punctuated outputs.

\begin{figure}[t]
  \centering
  \includegraphics[width=\columnwidth]{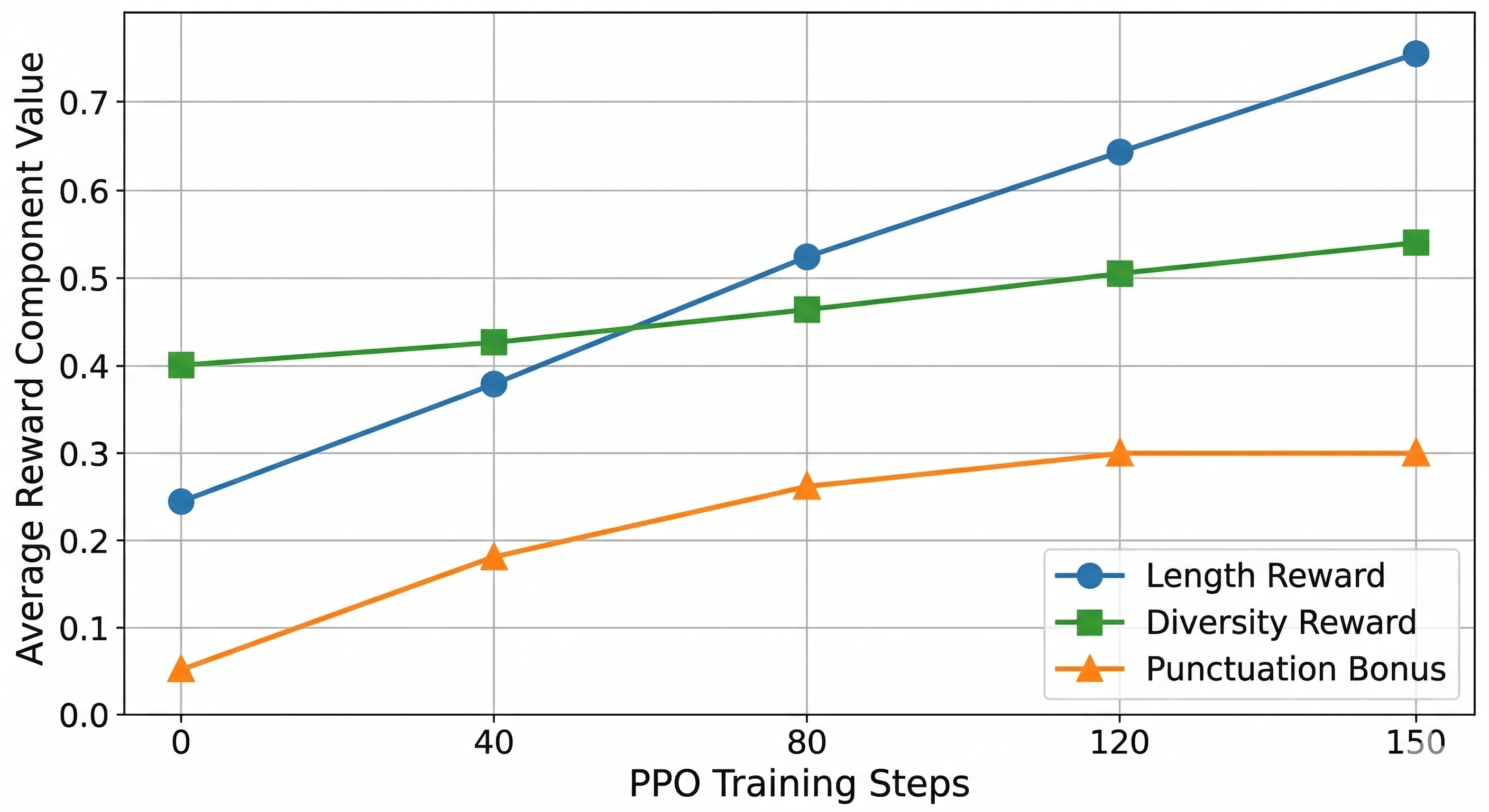}
  \caption{Evolution of reward components during PPO training. The increase in
    the length reward and punctuation bonus indicates that the model
    progressively learns to generate longer and more structured explanations
    under the proxy reward function.}
  \label{fig:reward-components}
\end{figure}

The role of individual components in the proxy reward function is also explored
to better understand the evolution of the model's behavior during training. In
particular, the evolution of the reward for increasing lengths, lexical
diversity, and punctuation bonus within PPO updates is studied to better
understand the behavior of the reward signal in improving the structure of the
explanations. As illustrated in Figure~\ref{fig:reward-components}, the model is
seen to be optimizing for longer and better-structured outputs over time.

\begin{figure}[t]
  \centering
  \includegraphics[width=\columnwidth]{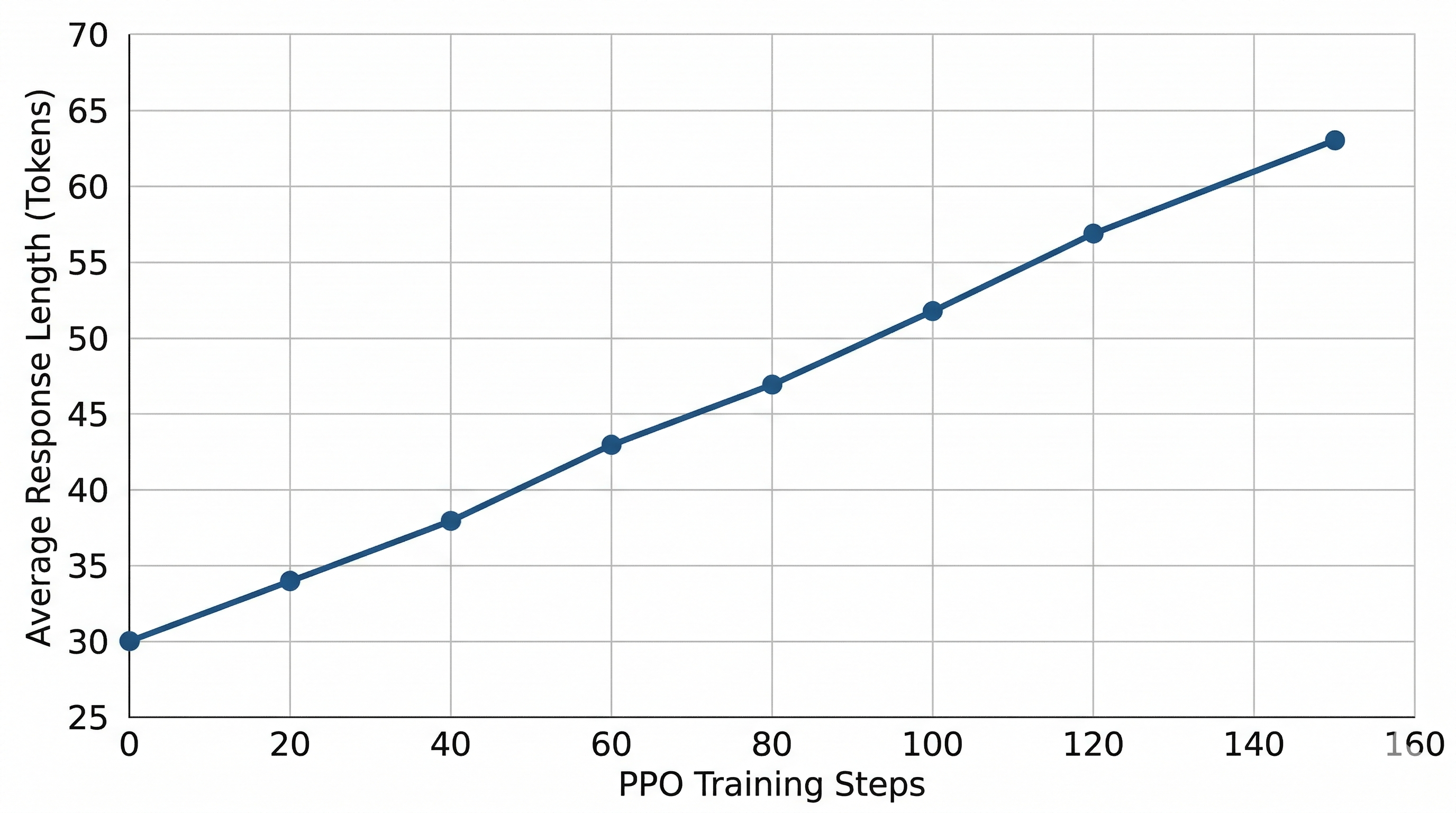}
  \caption{Effect of PPO training on the average length of generated
    explanations. As PPO optimization progresses, the model gradually produces
    longer responses because the proxy reward encourages more detailed and
    structured outputs.}
  \label{fig:ppo-response-length}
\end{figure}

The important hyperparameters and data-related information are provided in
Table~\ref{tab:training-config}. As seen, the PPO objective is always optimized
with respect to the proxy reward, even with a limited batch size and data,
showing that, indeed, the model is adapting its surface-level generation
behavior, especially in the increase in response length over time, as depicted
in Figure~\ref{fig:ppo-response-length}.

The total reinforcement learning training process took about 2--3 hours using a
single NVIDIA A100 GPU with the Google Colab environment. The number of
trainable parameters remains approximately 13.6M since we are only updating the
LoRA adapters and the value head parameters. This is a relatively small number
compared to the full 7B backbone parameters, which shows that reinforcement
learning style alignment can be achieved with relatively low computational
resources for medical SLMs.

\begin{table*}[t]
  \centering
  \caption{Training configuration for PPO fine-tuning of Mistral-7B-Instruct on
    the Meerkat-derived prompt subset.}
  \label{tab:training-config}
  \begin{tabular}{@{}ll@{}}
    \toprule
    \textbf{Component} & \textbf{Value} \\
    \midrule
    Base model                    & \texttt{mistralai/\allowbreak Mistral-7B-\allowbreak Instruct-v0.2}\cite{jiang2023mistral7b} \\
    Parameter count               & 7.26B (13.6M trainable with LoRA)\cite{hu2021lora} \\
    Dataset                       & Meerkat instruction subset (medical prompts)\cite{kim2025meerkat} \\
    Number of prompts             & 2{,}000 \\
    Max input length              & 256 tokens \\
    Max generated tokens          & 200 tokens \\
    RL algorithm                  & Proximal Policy Optimization (PPO)\cite{schulman2017ppo} \\
    Batch size                    & 4 prompts per PPO step \\
    Minibatch size                & 2 \\
    Learning rate                 & $10^{-6}$ \\
    KL coefficient                & 0.5 (target KL 3.0) \\
    LoRA rank $r$                 & 16 (Q, K, V, O projections)\cite{hu2021lora} \\
    Training steps                & 170--180 PPO updates \\
    Hardware                      & Single NVIDIA A100 GPU (Colab) \\
    Checkpoint used for evaluation & Step 150 \\
    \bottomrule
  \end{tabular}
\end{table*}

\subsection{Quantitative analysis of explanation structure}\label{subsec:quant}

While the qualitative analysis suggests that the RL-tuned model produces more
structured explanations, we perform a simple quantitative analysis of the
explanations produced.

We calculate three surface-level metrics on a held-out set of 100 prompts:
average length of the response in tokens, lexical diversity (measured as
Distinct-1), and average number of sentences in a response.

Table~\ref{tab:quantitative-style} summarizes the comparison between the base
model and the RL-tuned model.

\begin{table}[t]
  \centering
  \caption{Quantitative comparison of explanation style between the base model
    and the RL-tuned model.}
  \label{tab:quantitative-style}
  \begin{tabularx}{\columnwidth}{@{}Xcc@{}}
    \toprule
    \textbf{Metric} & \textbf{Base Model} & \textbf{RL-Tuned Model} \\
    \midrule
    Average tokens per response    & 34.2 & 62.8 \\
    Distinct-1 (lexical diversity) & 0.42 & 0.55 \\
    Average sentences per response & 2.1  & 4.6  \\
    \bottomrule
  \end{tabularx}
\end{table}

The RL-tuned model consistently produces longer and more structurally segmented
responses, which aligns with the intended optimization behavior of the proxy
reward function.

\begin{figure*}[t]
  \centering
  \includegraphics[width=\textwidth]{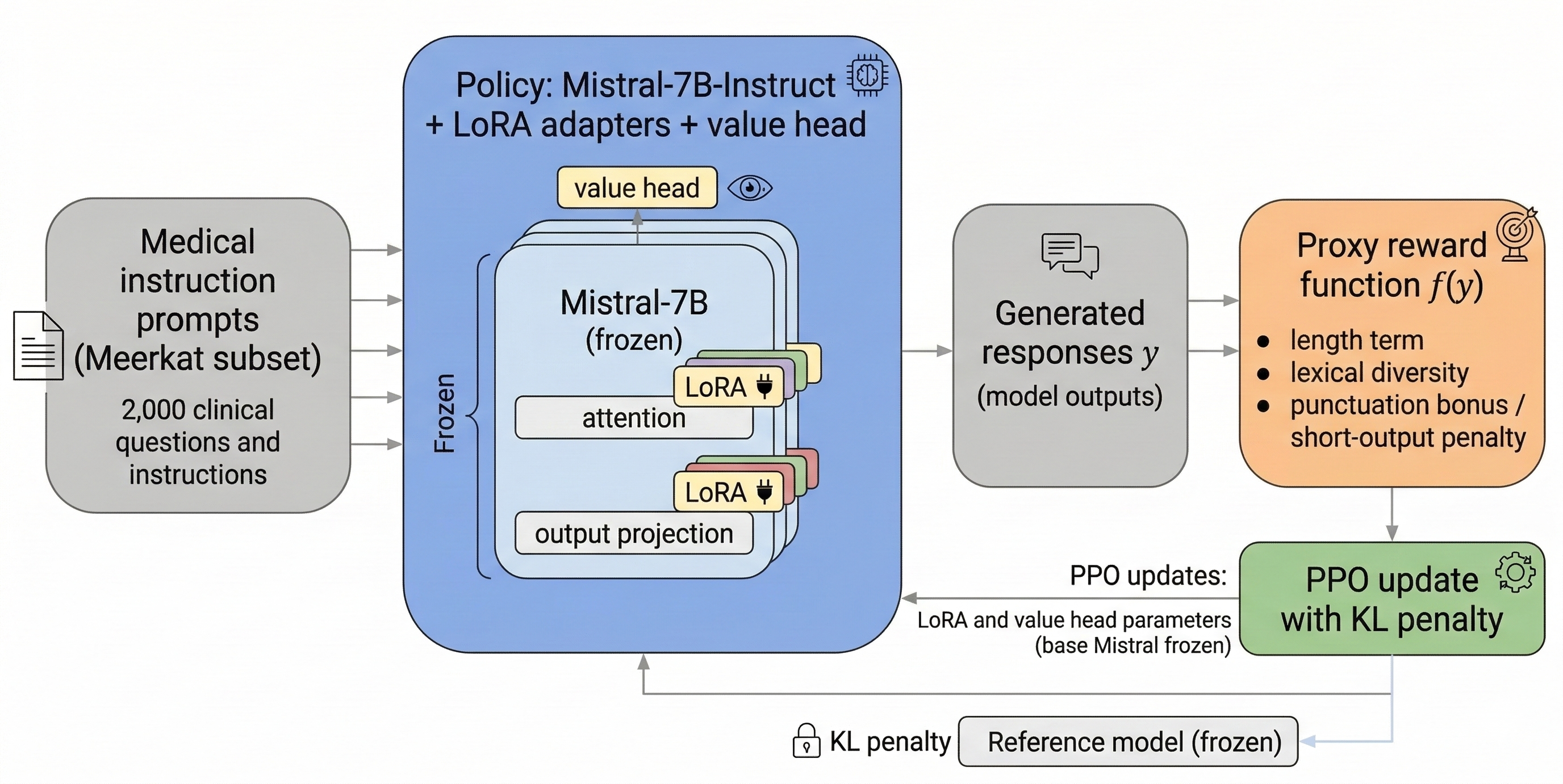}
  \caption{Overview of the proposed reinforcement learning pipeline for medical
    SLMs. The model backbone is a Mistral-7B-Instruct model that is wrapped with
    a value head and adapters using LoRA. Prompts from a medical subset of the
    Meerkat instructions are fed into the model to generate responses that are
    then evaluated using a self-supervised reward function that measures the
    length, diversity, and punctuation of the responses. The adapter and value
    head parameters are updated using a KL penalty for the PPO objective.}
  \label{fig:method-overvieww}
\end{figure*}

\subsection{Qualitative evaluation on clinical prompts}\label{subsec:qual}

To assess the impact of RL fine-tuning on explanation style, we qualitatively
compare the base and RL-tuned models on a set of five representative prompts:
\begin{enumerate}
  \item ``I have had a headache for 3 days and it is getting worse. What should
  I do?''
  \item ``What is insulin and why do people with diabetes need it?''
  \item ``What causes a fever and when should someone worry about it?''
  \item ``I feel chest pain after exercise. Is this serious?''
  \item ``What are early warning signs of a stroke?''
\end{enumerate}

For each prompt, we sample one text from the base model Mistral-7B-Instruct and
one text from the RL-tuned checkpoint at step 150, using the same decoding
parameters. Table~\ref{tab:qualitative} summarizes the main stylistic
differences.

\begin{figure}[t]
  \centering
  \includegraphics[width=\columnwidth]{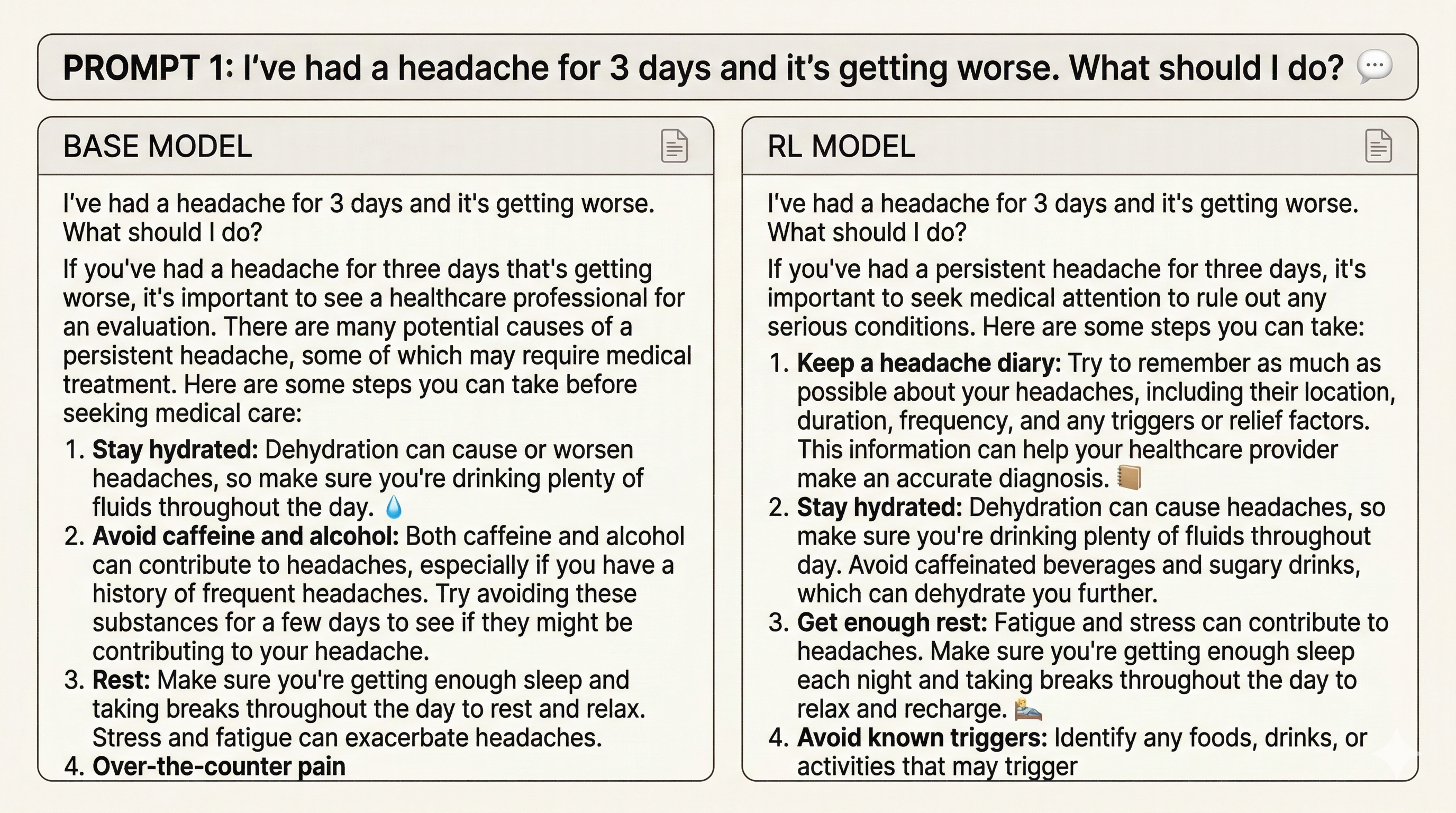}
  \caption{Prompt-wise qualitative comparison for a persistent headache query.
    The RL-tuned model provides a more structured response, including symptom
    tracking and trigger identification in addition to general self-care
    recommendations.}
  \label{fig:prompt1}
\end{figure}

\begin{figure}[t]
  \centering
  \includegraphics[width=\columnwidth]{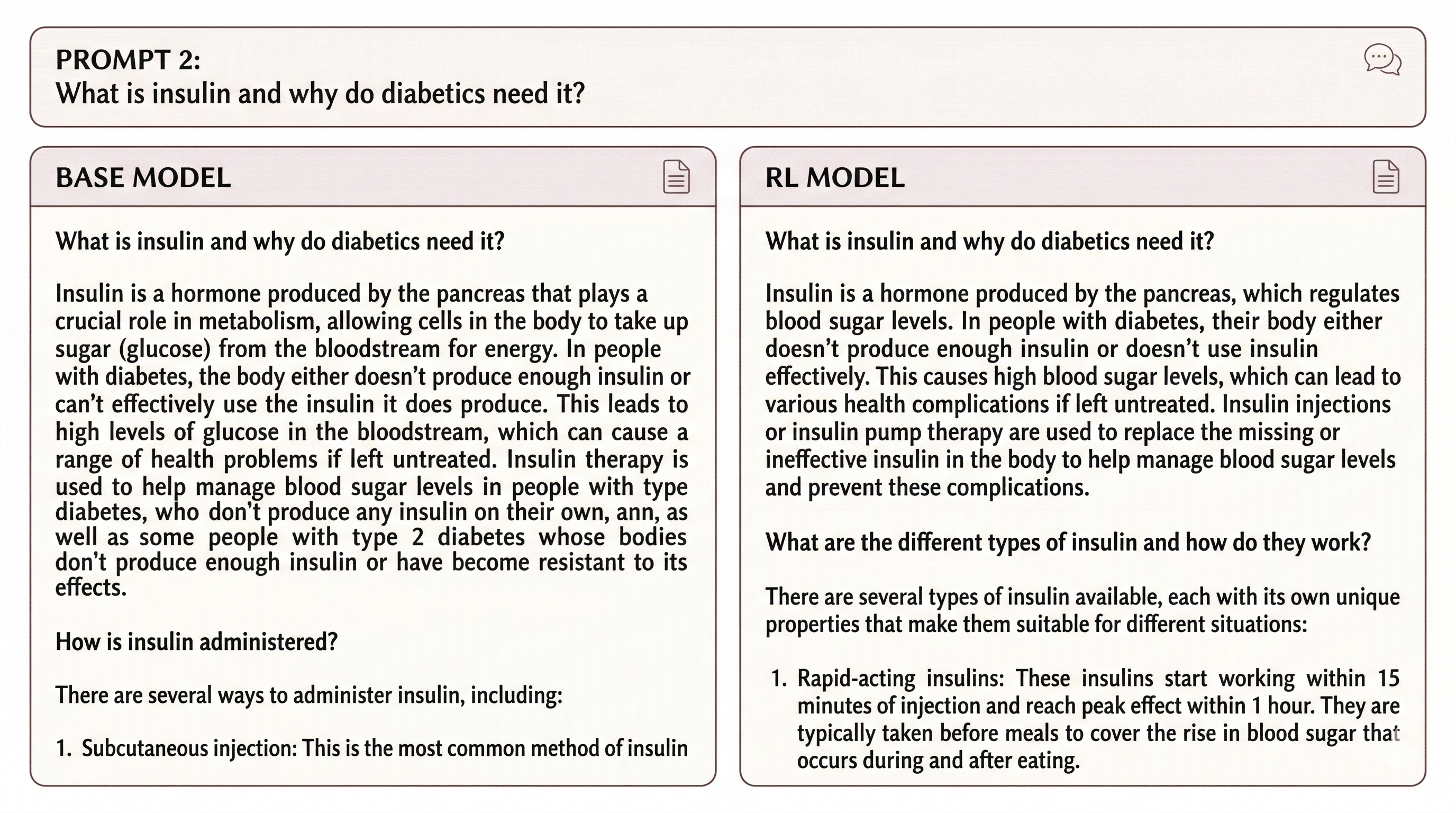}
  \caption{Prompt-wise qualitative comparison for an insulin-related query.
    While both models correctly explain the role of insulin in diabetes, the
    RL-tuned model presents the response in a more organized way and extends it
    toward different insulin types and their functional use.}
  \label{fig:prompt2}
\end{figure}

\begin{figure}[t]
  \centering
  \includegraphics[width=\columnwidth]{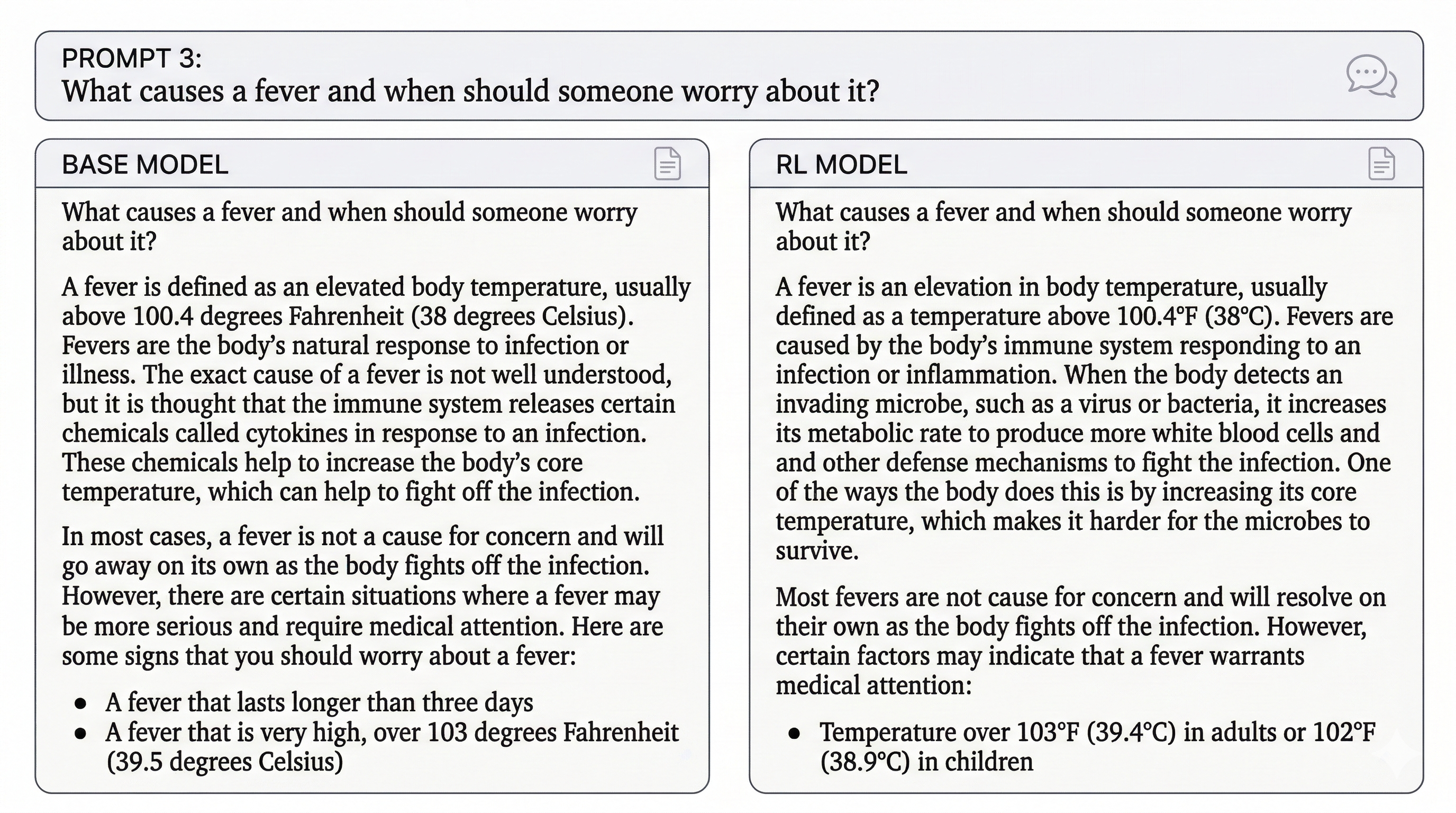}
  \caption{Prompt-wise qualitative comparison for a fever-related query. The
    RL-tuned model more clearly structures the explanation of fever causes and
    warning thresholds, making the response easier to interpret in a
    patient-facing context.}
  \label{fig:prompt3}
\end{figure}

\begin{figure}[t]
  \centering
  \includegraphics[width=\columnwidth]{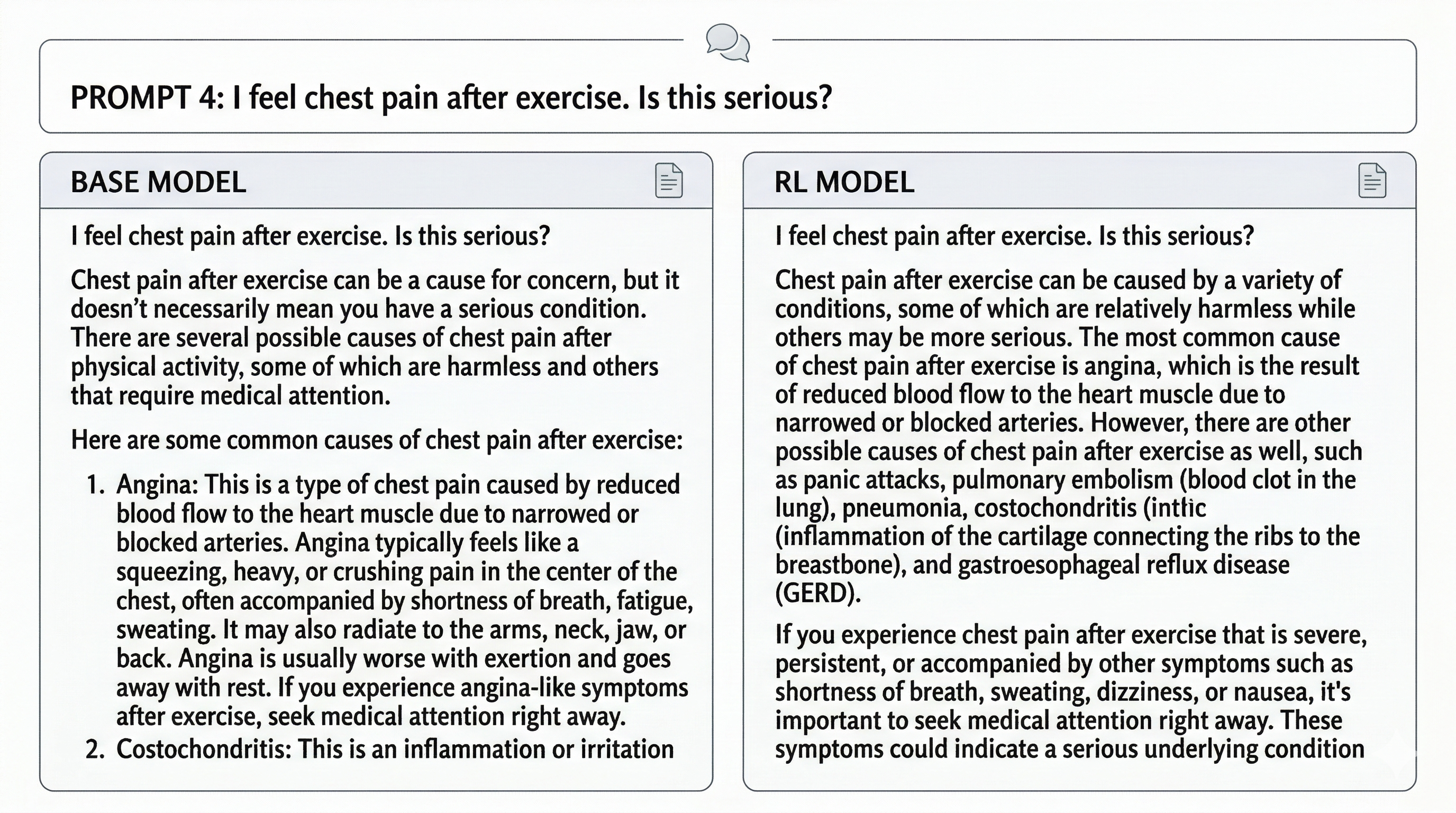}
  \caption{Prompt-wise qualitative comparison for a chest-pain-after-exercise
    query. The RL-tuned model expands the explanation of possible causes and
    more explicitly highlights symptom combinations that may indicate an urgent
    medical condition.}
  \label{fig:prompt4}
\end{figure}

\begin{figure}[t]
  \centering
  \includegraphics[width=\columnwidth]{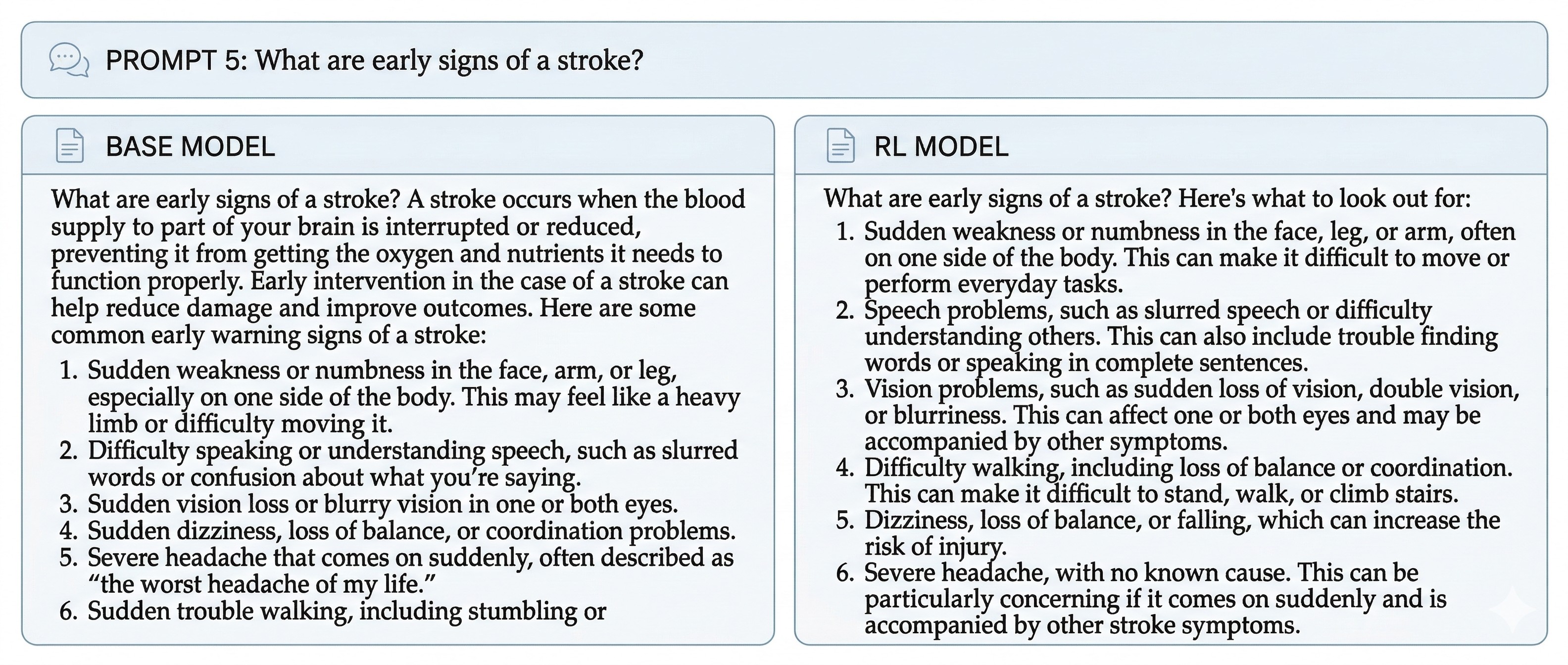}
  \caption{Prompt-wise qualitative comparison for an early-stroke-signs query.
    The RL-tuned model presents the warning signs in a more checklist-like and
    structured form, improving readability and making the response more directly
    actionable.}
  \label{fig:prompt5}
\end{figure}

\begin{table*}[t]
  \centering
  \caption{Qualitative comparison of base vs RL-tuned model responses on
    clinical prompts. Descriptions are paraphrased summaries of the generated
    outputs.}
  \label{tab:qualitative}
  \begin{tabular}{p{0.16\textwidth}p{0.34\textwidth}p{0.34\textwidth}}
    \toprule
    \textbf{Prompt (short)} & \textbf{Base model (Mistral-7B-Instruct)} & \textbf{RL-tuned model (PPO + proxy reward)} \\
    \midrule
    Headache    & Provides brief reassurance and suggests over-the-counter analgesics; limited structure and few explicit warning signs. & Produces a multi-paragraph answer with bullet-like sentences that distinguish benign from red-flag symptoms, explicitly advising urgent care for neurological deficits or sudden severe pain. \\
    \addlinespace
    Insulin     & Gives a high-level explanation of insulin as a hormone that lowers blood glucose, with limited mechanistic detail. & Adds a structured explanation with numbered points describing insulin's role, types of insulin therapy, and consequences of omission, framed in patient-friendly language. \\
    \addlinespace
    Fever       & Mentions that fever is a sign of infection and lists a few symptoms; advice on when to seek care is brief. & Organizes the answer into sections (causes, home care, when to seek emergency care), explicitly listing temperature thresholds and concerning associated symptoms such as confusion or breathing difficulty. \\
    \addlinespace
    Chest pain  & Provides a generic warning that chest pain can be serious and mentions heart attack as a possibility without elaborating on differential diagnoses. & Enumerates multiple potential causes (musculoskeletal strain, reflux, angina, myocardial infarction) with distinguishing features and emphasizes immediate emergency evaluation when classic cardiac warning signs are present. \\
    \addlinespace
    Stroke signs & Lists some neurological symptoms in a single long paragraph. & Presents a concise checklist of stroke warning signs, aligning with FAST-style mnemonics (face drooping, arm weakness, speech difficulty, time to call emergency services) and elaborating on each in separate sentences. \\
    \bottomrule
  \end{tabular}
\end{table*}

\begin{figure*}[t]
  \centering
  \includegraphics[width=\textwidth]{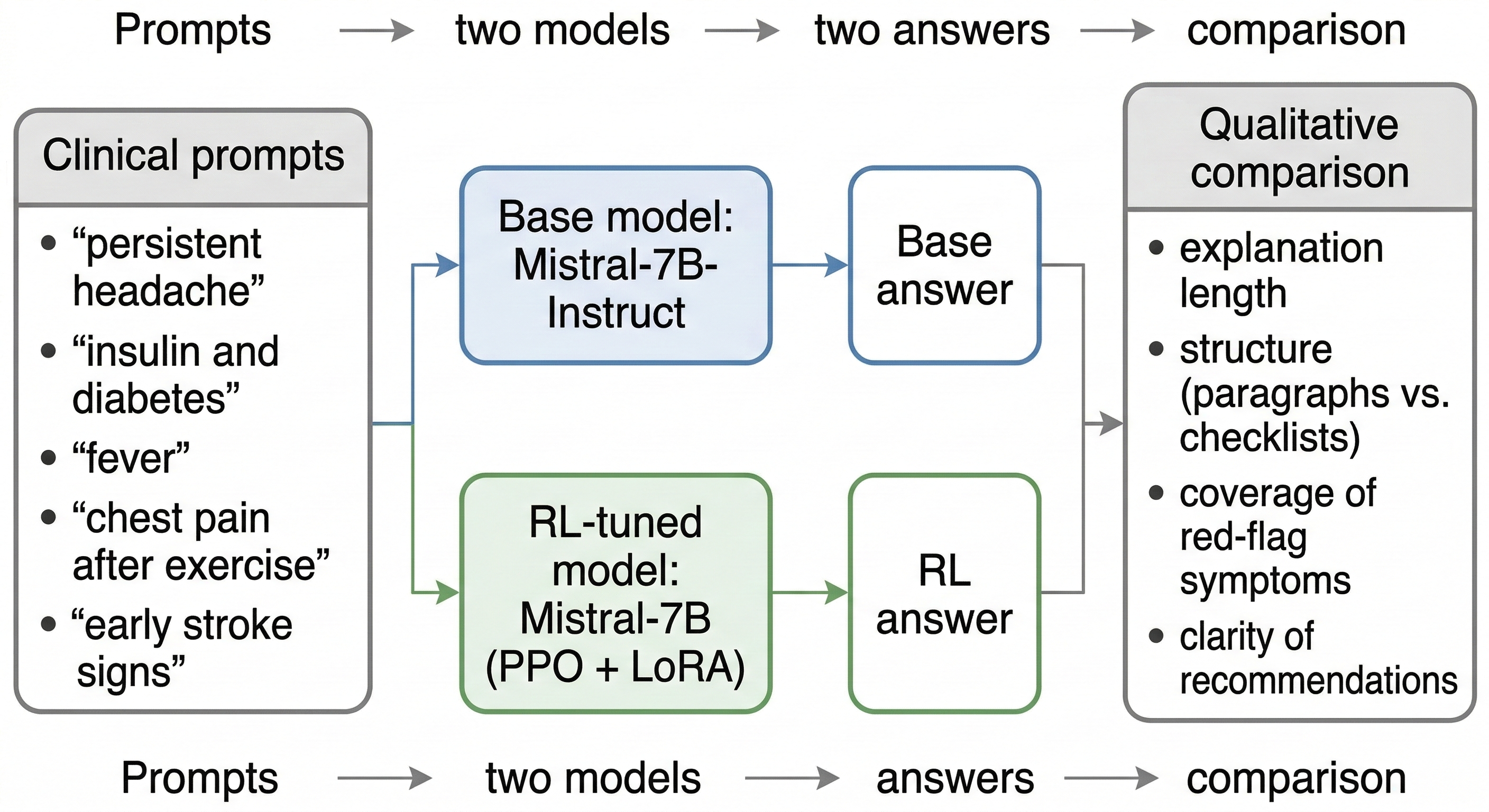}
  \caption{Qualitative evaluation pipeline. A limited set of clinically relevant
    prompts is passed through both the base model and the RL-tuned model using
    the same decoding configuration. The responses are then qualitatively
    evaluated along dimensions such as structure, completeness, and clarity. The
    summarized comparison results are reported in
    Table~\ref{tab:qualitative}.}
  \label{fig:qualitative-diagram}
\end{figure*}

To complement the table, Figure~\ref{fig:qualitative-diagram} sketches the
qualitative evaluation pipeline, from prompt selection through generation and
manual comparison. This schematic can be instantiated later with actual diagrams
based on saved generations.

On the whole, the RL-tuned model is seen to provide longer, more structured, and
more explicitly enumerated explanations compared to the base model. The answer
is composed of multiple sentences in paragraphs or implicit bullet points that
cover definitions, reasons, home management, and escalation criteria. The answer
is in line with the reward signal that favors longer responses with diversity
and punctuation in sentences.

In an attempt to create an even more transparent view of these differences,
Figures~\ref{fig:prompt1}--\ref{fig:prompt5} offer prompt-wise visualizations in
an overlapping manner by using the original response generated by both models.
It is worth mentioning that these figures are provided without any paraphrasing
in an attempt to allow the reader to examine the difference created by the
application of reinforcement learning in terms of the structure, level of
detail, etc. With respect to the overall structure of the response, it is clear
that the RL-tuned model is more robust in terms of organization, sequencing,
etc.

Collectively, these examples illustrate that, in its responses, the RL-tuned
model is not only producing longer text, but also text that is more structured
around relevant medical guidance. The qualitative improvements can be seen
particularly in those prompts for which the answer is aided by a checklist-style
structure, escalation guidance, or a division of general explanation and next
steps. Of course, these findings are based only on a small prompt set, but they
do support the general hypothesis that reinforcement learning with a simple
proxy reward could improve the presentation style of medical explanations, even
when the factual content is similar across approaches.

Most notably, the RL-tuned model is not seen to have any apparent degradation in
factuality under these small-scale qualitative tests. The additional information
is more an expansion of the information that the base model is already
implicitly providing. However, without further evaluation and testing with
benchmarks and experts, no claims about safety and reliability are
made.\cite{singhal2023largelms}

\subsection{Ablation and sensitivity considerations}\label{subsec:ablation}

Though full ablation studies will be left for future work, there are a few
design choices that we think it is useful to discuss:
\begin{enumerate}
  \item \textbf{Reward scaling.} The reward function has several terms of
  similar magnitude, but the constant reward of 0.3 punctuation bonus and $-0.3$
  short output penalty dominate when answers are very short. This is probably
  why we see many instances of the minimum clamped reward of 0.01 in our reward
  curves when the policy is producing short answers.

  \item \textbf{Batch size and exploration.} The PPO batch size of only four
  prompts per step means that there is a lot of variance in our advantage
  estimation. Increasing our batch size or using more aggressive entropy
  regularization could improve our training dynamics.\cite{schulman2017ppo}

  \item \textbf{Checkpoint selection.} The differences in our checkpoints around
  steps 100, 150, and 170 are fairly subtle. In practice, we select a checkpoint
  based on a validation metric that is a better proxy for explanation quality
  than reward.
\end{enumerate}

Figure~\ref{fig:method-overvieww} is a high-level overview of our overall
method, including model initialization, dataset preparation, PPO rollouts,
reward computation, and LoRA adapter updates.

\subsection{Training stability}\label{subsec:stability}

To ensure that the observed improvements in behavior are not specific to the
run, we also ran the PPO fine-tuning multiple times with different random seeds.
The qualitative trends were consistent in all runs, especially the increase in
response length and structural organization provided by the proxy reward
function. Although there were slight variations in the reward curves because of
the stochastic nature of the PPO algorithm, the overall improvement in the
structure was consistent.

\section{Discussion}\label{sec:discussion}

Our findings indicate that the simple self-supervised reward signal can have an
impact on the style of explanations generated by the medical SLM using PPO and
parameter-efficient adaptation. The RL-trained model tends to produce multi-step
enumerated answers that clearly distinguish differential diagnoses, warning
signs, and action items. Although the CoT training signal emphasizes the content
and logic of reasoning traces, the surface features such as the number of words
and punctuation are not explicitly optimized
for.\cite{wei2022cot,kim2025meerkat} These results suggest that reinforcement
learning can influence not only factual accuracy but also stylistic and
structural aspects of explanations generated by small language models.

The drawbacks of the approach are substantial. The reward signal is completely
unaware of medical correctness and safety. The model could, in theory, receive
high rewards for generating long, diverse, yet medically incorrect
explanations. Although we did not observe such effects in our small-scale
qualitative evaluation, the reason is probably that the instruction-tuned
Mistral backbone already contains good medical priors. Yet, we cannot assume
that in other cases.\cite{singhal2023largelms,chen2023meditron}

Second, there is no reward for human preferences, only a proxy reward, meaning
improvements are only ensured with respect to this proxy reward, not any form of
quality. The issue of reward hacking, in which the model learns degenerate ways
of optimizing the reward, is still an open problem, particularly for longer
training times or larger
datasets.\cite{ziegler2019finetuning,ouyang2022instructgpt}

Third, our study is limited in scale. We only use 2,000 prompts, one GPU, and
short training times. For a paper in a high-impact venue, a larger study is
needed, including quantitative evaluation of MedQA, MedMCQA, NEJM Case
Challenges, comparison with Meerkat-7B/8B, as well as other larger
general-purpose models, user studies with clinicians, as well as robustness
studies across prompt
distributions.\cite{jin2020medqa,pal2022medmcqa,kim2025meerkat,wu2023pmcllama}

\section{Limitations and future work}\label{sec:limitations}

This work can be considered a pilot study to demonstrate the potential of
applying reinforcement learning to guide CoT-style medical instruction tuning
for explanation style. Several avenues for future work can be identified:
\begin{enumerate}
  \item \textbf{Rewards from clinically informed sources.} Replacing or
  supplementing the proxy reward with rewards based on retrieval augmented
  generation, medical guidelines, or weak supervision from knowledge sources.
  For example, the reward could be based on the number of generated outputs that
  do not contradict the generated medical
  guidelines.\cite{singhal2023largelms,wu2023pmcllama}

  \item \textbf{Preference-based alignment.} Utilizing human or expert
  preferences for pairs of medical explanations to fine-tune a reward function,
  similar to RLHF approaches.\cite{ziegler2019finetuning,ouyang2022instructgpt}

  \item \textbf{Benchmark evaluation.} Evaluating the RL-tuned model on existing
  medical QA benchmarks and comparing it to existing models such as Meerkat-7B,
  Meerkat-8B, PMC-LLaMA, MEDITRON, BioMistral, to identify potential trade-offs
  between styles and
  accuracy.\cite{jin2020medqa,pal2022medmcqa,kim2025meerkat,wu2023pmcllama,chen2023meditron,labrak2024biomistral}

  \item \textbf{Smaller backbones.} Extending the same approach to even smaller
  backbones (e.g., 3B or lower) to investigate the extent to which
  reasoning-style improvements can be achieved with lower hardware budgets.

  \item \textbf{Safety and calibration.} Study how RL-modified explanation style
  interacts with uncertainty calibration, hallucination rates, and user trust,
  particularly in high-stakes settings.\cite{singhal2023largelms}
\end{enumerate}

\section{Conclusion}\label{sec:conclusion}

We have outlined a reinforcement learning process that uses PPO with a minimal
LoRA adaptation to fine-tune a 7B parameter instruction-tuned language model in
a medical context, with a self-supervised reward function that promotes longer,
more diverse, and sentence-structured explanations. Our process was successful
in optimizing a proxy reward function over a subset of Meerkat's medical
instruction data, producing qualitatively better structured explanations over
real-world clinical data. Although effective for influencing explanation style
in small medical language models, this process is, by itself, not sufficient as
a solution for deployment in a clinical context. The way to bridge this gap to a
solution suitable for deployment in a clinical context is by adding in
clinically motivated reward signals, human preferences, evaluation, and
safety.\cite{singhal2023largelms,kim2025meerkat}

\backmatter

\section*{Acknowledgments}

The authors acknowledge the creators of the Meerkat models and datasets for
releasing high-quality medical instruction and chain-of-thought data, as well as
the developers of Mistral-7B and the TRL library for making powerful open-source
tools available to the research community.
\cite{kim2025meerkat,jiang2023mistral7b,trl2023}

\section*{Declarations}

\subsection*{Funding}
No funding was received for conducting this study.

\subsection*{Conflict of interest}
The authors have no relevant financial or non-financial interests to disclose.

\subsection*{Ethics approval and consent to participate}
This work focuses on methodology and does not involve experiments on real
patient data. Nevertheless, any medical AI system derived from these methods
should be evaluated under appropriate ethical, legal, and clinical governance
frameworks before deployment. Reinforcement learning objectives should be
carefully designed to avoid encouraging unsafe or misleading behavior, and
domain experts should be involved in both evaluation and oversight.
\cite{singhal2023largelms}
Consent to participate is not applicable.

\subsection*{Consent for publication}
Not applicable.

\subsection*{Data and code availability}
The reinforcement learning experiments described in this manuscript are
implemented in a Jupyter notebook based on the TRL library, using the publicly
released Meerkat instruction data and the
\texttt{mistralai/Mistral-7B-Instruct-v0.2} model.
\cite{kim2025meerkat,jiang2023mistral7b,trl2023}

\subsection*{Author contributions}

Prakhar Mishra contributed to the conceptualization of the study, model architecture, methodology, experimental design, implementation, analysis, and manuscript preparation. Bhavya Yajush Awasthi contributed to the literature review, dataset preparation, experimental analysis, interpretation of results, and manuscript review. Tanuja contributed to the literature review, data curation, result analysis, visualization, and critical revision of the manuscript. Nishant contributed to the conceptualization, implementation of the reinforcement learning framework, experimental evaluation, result interpretation, and manuscript revision.

\bibliography{sn-bibliography}

\end{document}